\documentclass[conference]{IEEEtran}
\IEEEoverridecommandlockouts
\usepackage{cite}
\usepackage{amsmath,amssymb,amsfonts}
\usepackage{algorithmic}
\usepackage{graphicx}
\usepackage{textcomp}
\usepackage{xcolor}
\usepackage{subfigure}
\def\BibTeX{{\rm B\kern-.05em{\sc i\kern-.025em b}\kern-.08em
    T\kern-.1667em\lower.7ex\hbox{E}\kern-.125emX}}
\begin{document}

\title{Bokeh Rendering Based on Adaptive Depth Calibration Network}

\author{\IEEEauthorblockN{Lu Liu}
\IEEEauthorblockA{\textit{Shenzhen International Graduate School} \\
\textit{Tsinghua University}\\
Shenzhen, China \\
l-liu20@mails.tsinghua.edu.cn}
\and
\IEEEauthorblockN{Lei Zhou}
\IEEEauthorblockA{\textit{Independent Researcher} \\
leizhou.astro@gmail.com}
\and
\IEEEauthorblockN{\textsuperscript{*}Yuhan Dong}
\IEEEauthorblockA{\textit{Shenzhen International Graduate School} \\
\textit{Tsinghua University}\\
Shenzhen, China \\
dongyuhan@sz.tsinghua.edu.cn}
}

\maketitle

\begin{abstract}
Bokeh rendering is a popular and effective technique used in photography to create an aesthetically pleasing effect.
  It is widely used to blur the background and highlight the subject in the foreground, thereby drawing the viewer's attention to the main focus of the image. 
  In traditional digital single-lens reflex cameras (DSLRs), this effect is achieved through the use of a large aperture lens. 
  This allows the camera to capture images with shallow depth-of-field, in which only a small area of the image is in sharp focus, while the rest of the image is blurred. 
  However, the hardware embedded in mobile phones is typically much smaller and more limited than that found in DSLRs. 
  Consequently, mobile phones are not able to capture natural shallow depth-of-field photos, which can be a significant limitation for mobile photography. 
  To address this challenge, in this paper, we propose a novel method for bokeh rendering using the Vision Transformer, a recent and powerful deep learning architecture. 
  Our approach employs an adaptive depth calibration network that acts as a confidence level to compensate for errors in monocular depth estimation. This network is used to supervise the rendering process in conjunction with depth information, allowing for the generation of high-quality bokeh images at high resolutions. 
  Our experiments demonstrate that our proposed method outperforms state-of-the-art methods, achieving about 24.7\% improvements on LPIPS and obtaining higher PSNR scores.
\end{abstract}

\section{Introduction}
\begin{figure}[htbp]
\centerline{\includegraphics[width=9cm]{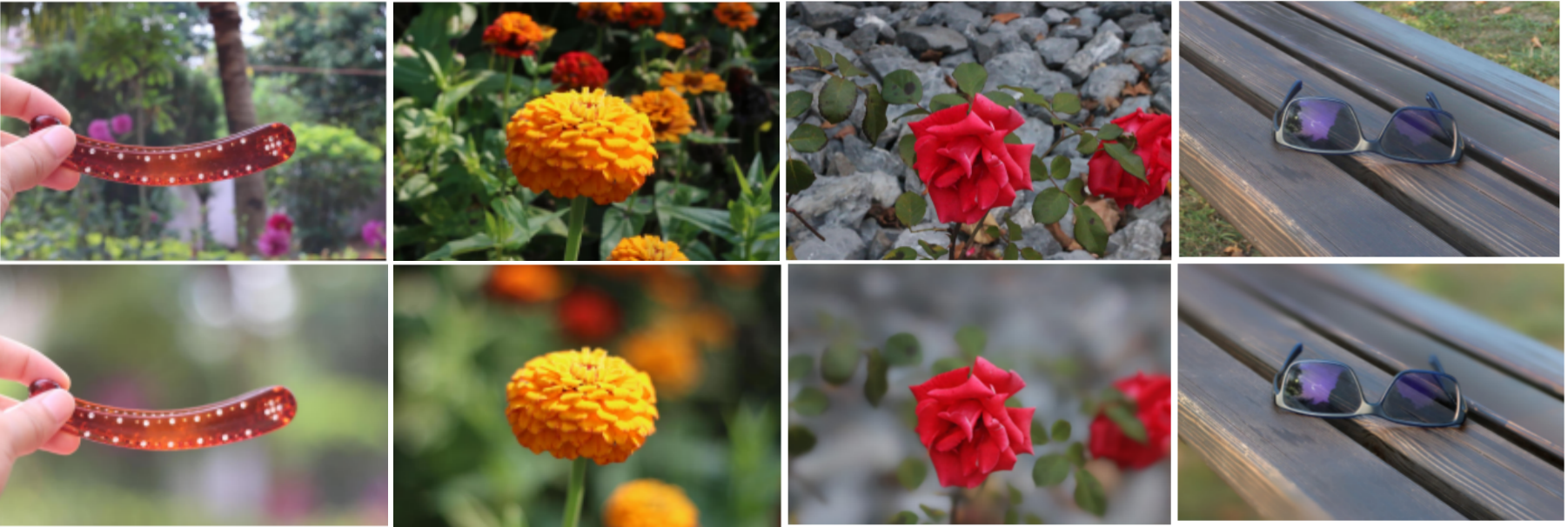}}
\caption{Example images of the proposed bokeh rendering effect. The top row shows the all-in-focus images, while the bottom row shows the bokeh rendering effect applied to the same scenes.}
\label{H}
\end{figure}

Bokeh rendering is a popular artistic effect in photography that enhances the perception of an image's foreground and background, resulting in visually appealing effects (see Fig. \ref{H}). This technique is achieved by creating a shallow depth of field that blurs the background and highlights the subject in the foreground. While bokeh rendering has been traditionally achieved through the use of specialized lenses, recent advancements in computer vision have led to the development of synthetic bokeh rendering techniques.
Early works in synthetic bokeh rendering were limited to portrait mode and were not adaptable to other photography scenes \cite{shen2016automatic, shen2016deep}. In recent years, several computer vision models \cite{ignatov2020rendering, nagasubramaniam2022bokeh, lee2022bokeh} have shown excellent results in bokeh rendering tasks. However, these models have not incorporated accurate prior depth information, which often results in unstable and unnatural transitions between the foreground and background.
To address this issue, this paper proposes a hybrid framework called BRADCN for achieving stable and realistic bokeh effects. The proposed framework leverages VIT \cite{dosovitskiy2020image} as the fundamental bokeh framework, allowing it to retain feature maps of the global receptive field. Additionally, Vision Transformer for Dense Predictions (DPT) \cite{ranftl2021vision} is applied to obtain a depth map that incorporates feature maps as input for the decoder. Due to the limited accuracy of monocular depth estimation, the paper introduces an Adaptive Depth Calibration Net (ADCN) to adapt to the error in depth estimation. The ADCN is trained on a real-world RGB-D dataset and is capable of providing realistic confidence for depth. The proposed method achieves state-of-the-art results in bokeh rendering tasks, improving evaluation metrics and producing significantly better visual effects.

The contributions of this work are are summarized as follows:
\begin{itemize}
    \item We propose a novel hybrid framework called BRADCN for achieving stable and realistic bokeh effects in synthetic bokeh rendering. 
    This framework integrates the Vision Transformer (VIT) and Vision Transformer for Dense Predictions (DPT) to retain feature maps and obtain a depth map, respectively.
    \item Incorporation of accurate prior depth information: The proposed framework addresses the issue of unstable and unnatural transitions between foreground and background in bokeh rendering by incorporating accurate prior depth information. 
    \item The proposed method achieves state-of-the-art results in bokeh rendering tasks, improving evaluation metrics and producing significantly better visual effects, which is a significant contribution to the field.
\end{itemize}

\section{RELATED WORK}

\subsection{Classical rendering}
Classical rendering techniques are a popular method for achieving bokeh effects in photography. These techniques can be broadly classified into two categories, namely, ray tracing and depth map construction.
The ray-tracing \cite{abadie2018advances} based method is computationally intensive and involves the reconstruction of 3D information in the light field. As a result, this method is not very practical for use in engineering applications.
The other classical method for bokeh rendering relies on depth information \cite{luo2020bokeh}. However, this method has not shown satisfactory results, primarily due to the challenge of accurately estimating depth information. The accuracy of monocular depth estimation is difficult to achieve in both academia and industry, and this poses a significant challenge for bokeh rendering. If the depth information is not accurate enough, the effectiveness of the bokeh rendering effect will be reduced.

\subsection{Rendering Based on Deep Learning}
With the advancement of computing hardware, deep learning-based methods have emerged as the preferred approach for bokeh rendering tasks. 
In particular, the work by \cite{ignatov2020rendering} achieved a realistic bokeh effect through an end-to-end manner that processes the input at multiple scales. 
Building upon this, [4] proposed a method based on VIT \cite{nagasubramaniam2022bokeh} and introduced an input-to-input mapping to better retain the details of the foreground. 
Another study by \cite{lee2022bokeh} explored the application of generative adversarial networks (GANs) to the rendering task, which provided a new approach for bokeh rendering tasks. 
\cite{lijun2018deeplens} proposed a rendering network that adopted a multitasking mechanism, incorporating depth prediction and semantic segmentation to achieve controllable focal distance and aperture size. Similarly, \cite{peng2022bokehme} integrated classical and neural rendering methods, using a neural network to adapt the error of classical rendering. 

Despite the progress made by these prior works, accurate depth information has not been fully integrated into the deep learning-based methods, resulting in decreased bokeh rendering effects to varying degrees. 
As such, the need for more sophisticated and precise depth estimation methods has been recognized to improve the performance of deep learning-based bokeh rendering techniques.

\section{PROPOSED METHOD}
\begin{figure*}[htbp]
\centerline{\includegraphics[width=\textwidth]{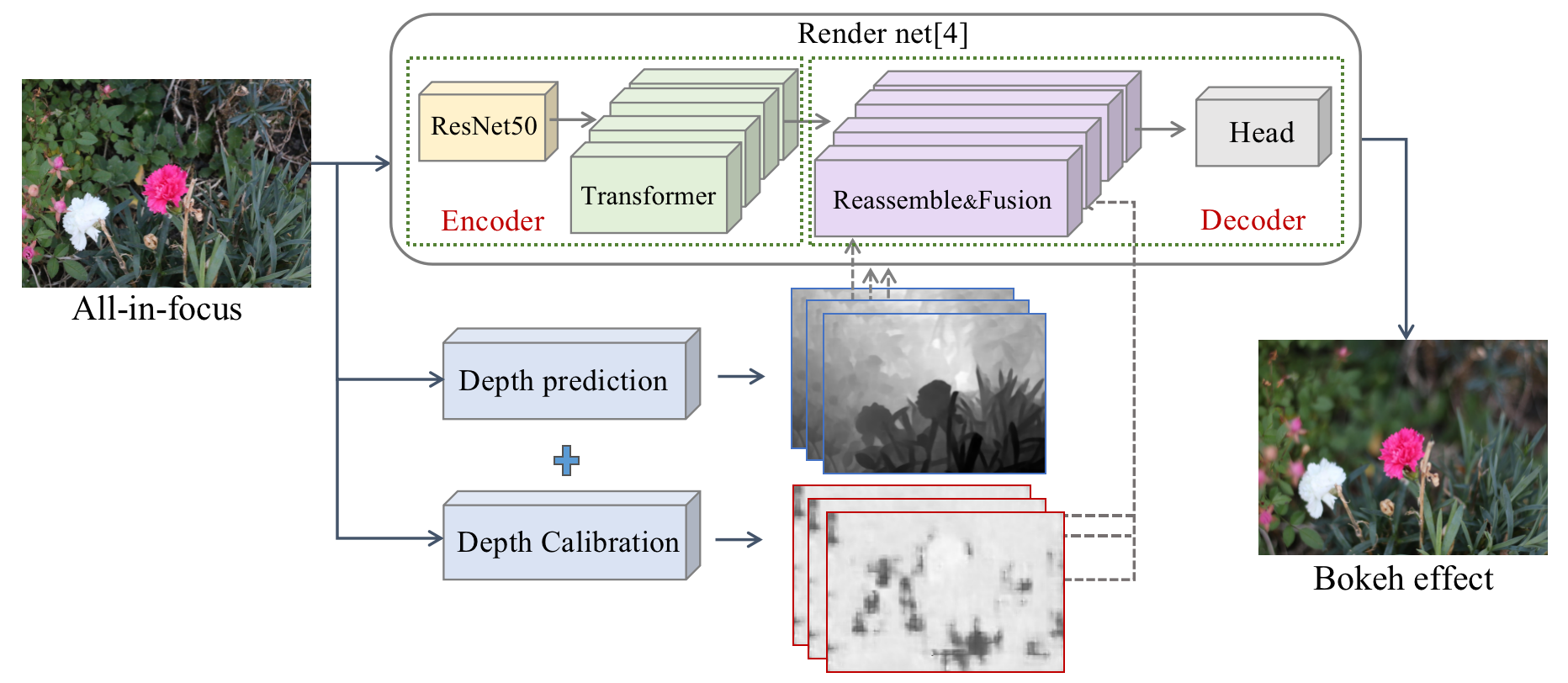}}
\caption{The overall architecture of the proposed BRADCN framework is designed to achieve the bokeh rendering effect in a data-driven manner. It consists of three sub-models, namely the Render Net, Depth Prediction Net, and Adaptive Depth Calibration Net.}
\label{M}
\end{figure*}
In Fig. \ref{M}, we present the architecture of our proposed framework, named BRADCN, which comprises of three primary sub-models: the Render Net, Depth Prediction Net, and Adaptive Depth Calibration Net. The Render Net is responsible for rendering synthetic images, while the Depth Prediction Net predicts the depth map of the input image. The Adaptive Depth Calibration Net is used to refine the predicted depth maps by calibrating the depth values according to the relative depth of the objects in the scene.

\subsection{Render Net}

In this section, we present the details of the Render Net, which is the first sub-model of our proposed framework. The Render Net is designed to generate synthetic images that are used to realize original bokeh rendering. We base our Render Net on the BRVIT architecture proposed in \cite{nagasubramaniam2022bokeh}, but we make some improvements and innovations to better suit our specific task.

First, we use a ResNet-50 as the backbone to extract features from the input images, as it is a widely used and effective approach in image processing. The ResNet-50 is a deep convolutional neural network that can extract high-level features from the input images. Next, the feature maps extracted from the ResNet-50 are divided into non-overlapping patches and flattened into tokens. These tokens, along with position embeddings and class tokens, are fed into the transformer blocks, which are the key components of the BRVIT architecture. The transformer blocks use self-attention mechanisms to model the relationships between different tokens and capture the global context of the image. The outputs of the transformer blocks are reassembled into different resolutions that contain information at multiple scales. To achieve this, we use fusion blocks, which combine the features from different resolutions to produce high-quality feature maps that contain multi-scale information. The fusion blocks use skip connections to progressively upsample low-resolution features and combine them with high-resolution features.
Finally, we apply a convolutional layer to fuse the outputs of the fusion blocks to the same resolution as the input images. The convolutional layer further processes the features to generate high-quality synthetic images that are visually similar to real images.

Overall, our Render Net improves on the BRVIT architecture by introducing fusion blocks that enable the network to capture multi-scale information and generate high-quality synthetic images. The ResNet-50 backbone, transformer blocks, and convolutional layer are standard components in computer vision tasks, and our use of these components in the Render Net is a well-established approach to feature extraction and image synthesis.

\subsection{Depth Prediction Net}

In our framework, we use the Depth Prediction Net to estimate the depth maps of the scene. Specifically, we apply the Dense Prediction Transformer (DPT) \cite{ranftl2021vision}, which is based on the Vision Transformer, to generate depth maps during both the training and inference stages. DPT is known for its state-of-the-art performance in dense prediction tasks.
We utilize the pre-trained model which is finetuned on the NYUv2 dataset.
To ensure that our model is able to handle the complexities of real-world scenes, we also account for differences in the semantic distribution of different datasets and fine-tune the weights during the training stages.
After generating the depth map using DPT, we apply an adaptive net to match the depth information with the multi-channel RGB information in the images. This adaptive net includes a convolutional layer with an output channel of 3, kernel size of 3, stride of 1, and padding of 1. The output of the adaptive net is reassembled into different resolutions as mentioned in Section 3.1.
To combine the information from the encoder of the Render Net and the Depth Prediction Net, we use the decoder input defined as:
\begin{equation}\label{e1}
\mathit{D}{i} = \mathit{E}{o} + \mathbf{Reassemble}(\mathbf{Adap}(\mathit{D}{o}))
\end{equation}
where \textit{Eo} is the output of the encoder of the Render Net and \textit{Do} is the output of the Depth Prediction Net after being processed by the adaptive net and reassembled into different resolutions.

\subsection{Adaptive Depth Calibration Net}

In this section, we describe the Adaptive Depth Calibration Net (ADCN), which is used to adapt the results of the depth prediction network to improve depth estimation accuracy. 
We note that monocular depth estimation algorithms have limitations in accuracy, and therefore, ADCN is introduced to calibrate the depth estimation results.
The ADCN was trained using the SUNRGBD dataset \cite{song2015sun}, a real-world indoor scene dataset. 
Generally, selecting a dataset with consistent semantic information with bokeh rendering dataset for training the ADCN model is a valid approach to ensuring that the model is able to handle the complexities of real-world scenes. 
By training the model on a dataset with similar semantic information, the model is able to learn features and patterns that are relevant to the target task.

However, it is important to note that the model may not generalize well to other datasets that have different semantic distributions. This means that the model may perform poorly on datasets that have a different semantic structure, even if they are related to the target task. Therefore, it is important to evaluate the model on a diverse set of datasets to ensure that it is able to generalize well to new scenes.

If the ADCN model is able to perform well despite the differences in semantic information, it suggests that the model has learned to extract features that are relevant to the target task regardless of the semantic information. This could be an indication of the model's ability to generalize to a broader range of real-world scenes.

\begin{figure}[ht]
\centerline{\includegraphics[width=9cm]{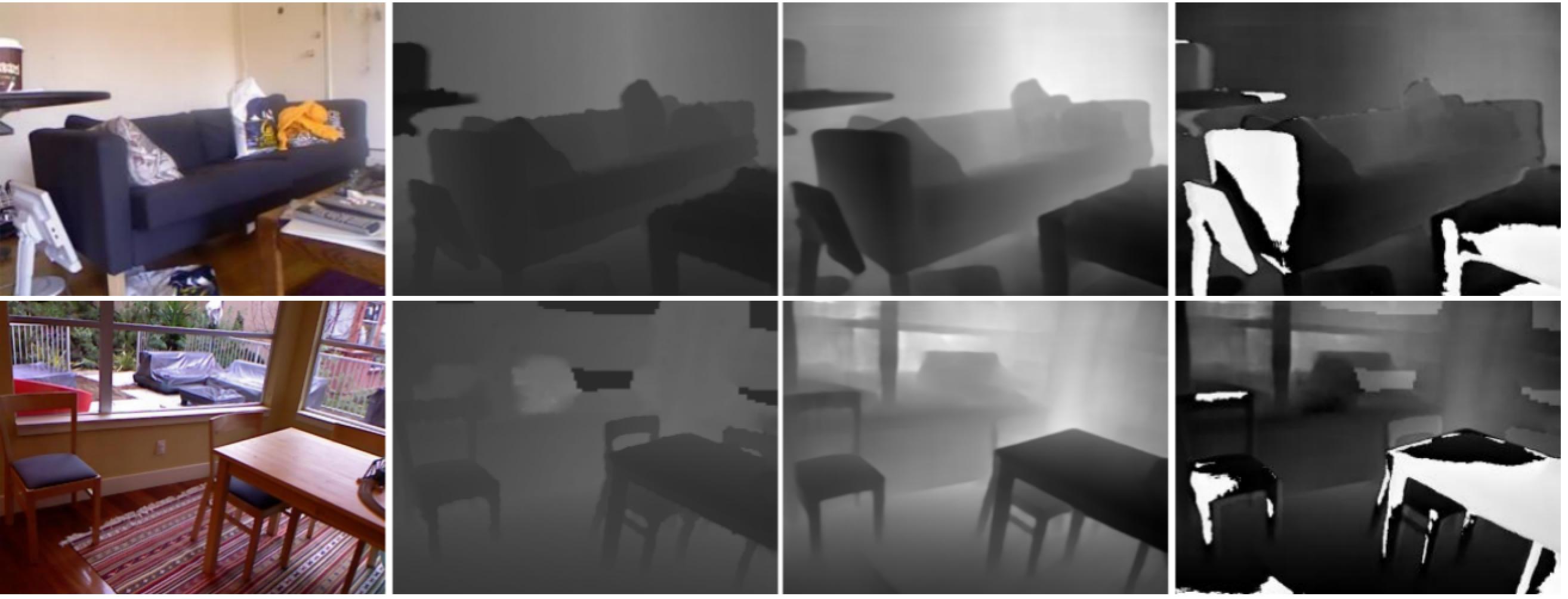}}
\caption{Visualization of obtaining the dataset for ADCN involves several stages. The input images, which are captured using a camera, are shown in the leftmost column. The \textit{Do} images, which are obtained by Depth Prediction Net, are shown in the second column. The \textit{D\_{gt}} images, which represent the ground truth depth maps, are shown in the third column. The last column shows the \textit{De} images, which are the error depth maps.}
\label{A}
\end{figure}

For SUNRGBD, our selection of the training set, validation set, and test set is completely random.
First, we feed the images from the dataset to the depth prediction network to predict the depth with the same weight as mentioned in Section 3.2. Then, we calculate the absolute error map between the predicted depth and the ground-truth depth in the SUNRGBD dataset. The absolute error map is used as the supervising data for the ADCN. It is described as :
\begin{equation}
\mathit{D}{e} =| \mathit{D}{o} - \mathit{D}\_{gt} |
\end{equation}
where \textit{Do} is the output of depth prediction, \textit{D\_{gt}} is the ground truth of depth. Thus, we obtain the supervising data of ADCN, namely the label. The process is visualized in Fig. \ref{A}.

\begin{figure}[htbp]
\centerline{\includegraphics[height=5cm]{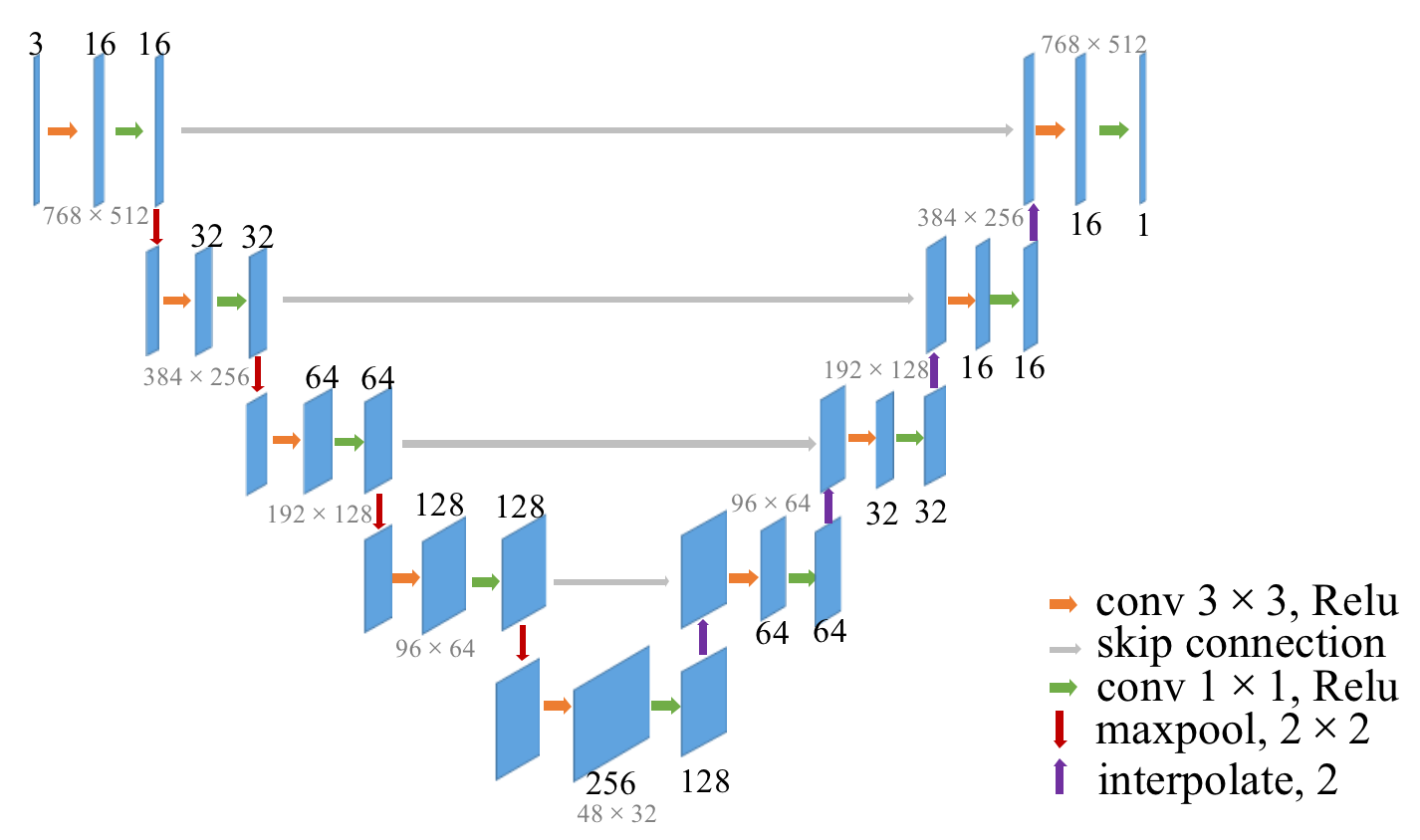}}
\caption{The overall architecture of ADCN.
The left side is an encoder which contains feature extraction and downsampling and the right side is a decoder for upsampling. Skip-connections are utilized between the encoder and decoder to capture multi-scale features.}
\label{U}
\end{figure}

The architecture of ADCN is based on U-Net \cite{ronneberger2015u}, which has outstanding matching ability and fast convergence speed. We show the architecture of ADCN in Fig. \ref{U}. The outputs of ADCN are fed into an adaptive net to match depth error information with multi-channel RGB information. The adaptive depth error is reassembled into different resolutions as mentioned in Section 3.1. The reassembled adapted error information and the \textit{Di} in Eq. (\ref{e1}) are added together as decoder input, which is defined as:
\begin{equation}
\mathit{D}{I} = \mathit{D}{i} + \mathbf{Reassemble}(\mathbf{Adap}(\mathit{D}{e}))
\end{equation}
where \textit{DI} is the input of the decoder of render net, \textit{Di} is mentioned in Eq. (\ref{e1}).


\section{EXPERIMENT}

In this section, we provide a detailed description of the datasets and loss functions used in our experiments. Furthermore, we compare the evaluation metrics of our proposed framework with state-of-the-art works in the field of bokeh rendering. Lastly, we present the results of some ablation experiments to demonstrate the effectiveness of the components in our framework.

\subsection{Dataset}

\textbf{EBB!} The EBB! dataset \cite{ignatov2020rendering}), short for "Everything is Better with Bokeh!", is a collection of over 5,000 image pairs that were captured in an outdoor scene using a Canon 7D DSLR camera. The image pairs in EBB! consist of an all-in-focus image captured using an aperture of f/16 and a bokeh-effect image captured using a wide aperture of f/1.8. Each image pair has a resolution of 1024 x 1536. For this study, 4400 image pairs were selected for training and 294 pairs were used for testing, following the same protocol as in \cite{nagasubramaniam2022bokeh}.

\textbf{SUNRGBD} The SUNRGBD dataset \cite{song2015sun} was published by the Vision Robotics Group of Princeton University and contains more than 10,000 images with corresponding depth information. For training the Adaptive Depth Calibration Network (ADCN), 6545 image pairs were choosed randomly, and 700 pairs were used for testing to enable the ADCN to adapt the depth error properly. The weights of the trained ADCN were saved for subsequent rendering tasks.

\subsection{Loss Functions}
In this work, we employ two loss functions for the two training stages. The first stage involves the pre-training of ADCN, and we use the L1 loss function \cite{zhao2016loss} for 50 epochs with a learning rate of 1e-4. For the second stage of training, which involves training the whole architecture, we initially train the hybrid model with the L1 loss function for 30 epochs until the loss function decreases smoothly. Then, we combine the L1 loss function with the MS-SSIM loss function \cite{wang2003multiscale} for an additional 30 epochs. The learning rate during this stage is set to 5e-5. The L1 loss function computes the mean absolute difference between the predicted and ground truth depth maps, whereas the MS-SSIM loss function measures the structural similarity between the predicted and ground truth depth maps at multiple scales. The combination of these two loss functions ensures that the model generates bokeh rendering effect that is visually pleasing.

\begin{figure*}[ht!]
  \centering
    \begin{subfigure}
      \centering   
      \includegraphics[width=0.98\linewidth]{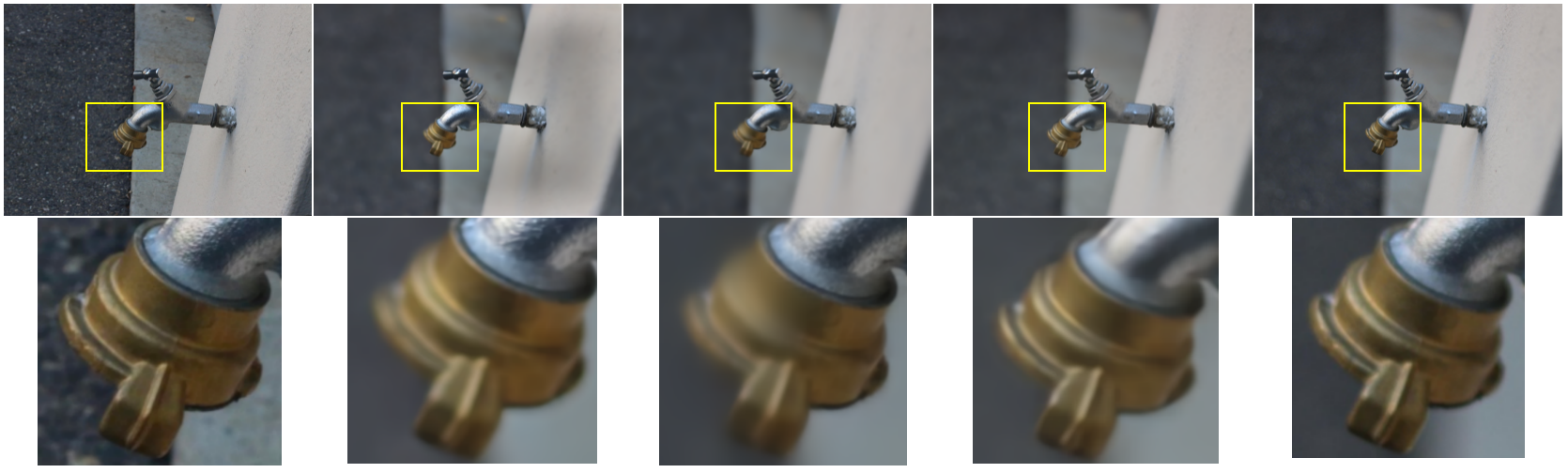}
    \end{subfigure}  
    \begin{subfigure}
      \centering   
      \includegraphics[width=0.98\linewidth]{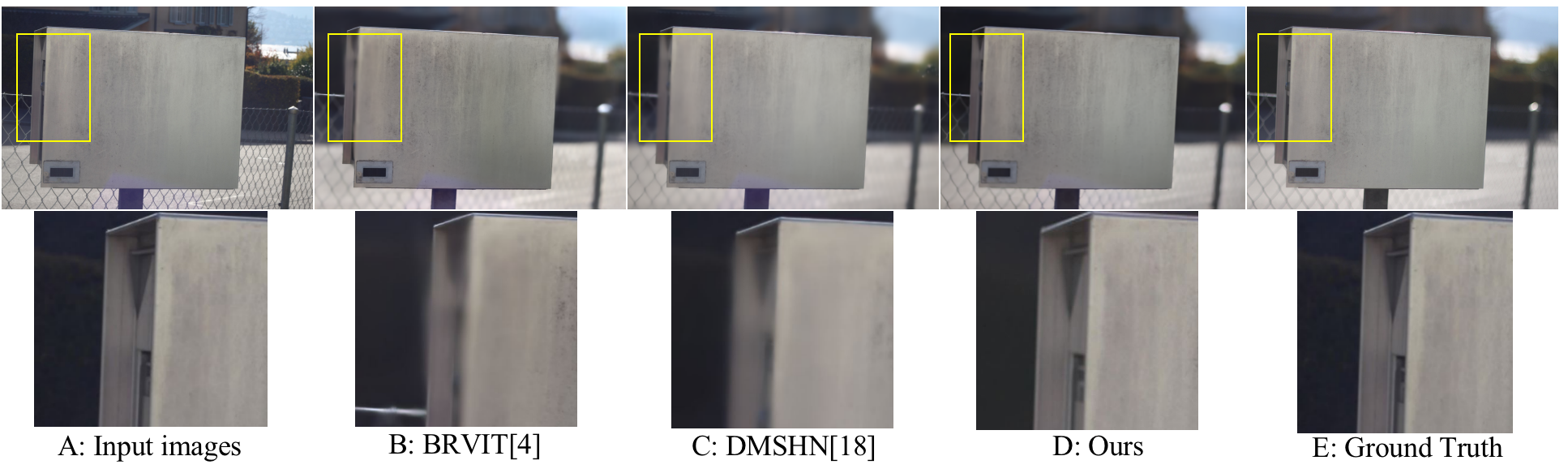}   
    \end{subfigure}
\caption{
\label{V}
Qualitative results comparison on the EBB! Val294 dataset. Our proposed model is able to create a more natural and smooth bokeh rendering effect, resulting in a sharper foreground and blurred background, as shown in the example images.}
\end{figure*}

\subsection{Metrics}
We apply three common objective metrics in image quality evaluation to compare with state-of-the-art works to demonstrate the effect of our model jointly, they are PSNR \cite{paul2019analysis}, SSIM \cite{zhao2016loss} and LPIPS \cite{zhang2018unreasonable}.
\begin{itemize}
    \item Peak Signal to Noise Ratio (PSNR) measures the difference between two images by comparing their peak signal power to the mean squared error (MSE) between them. 
    A higher PSNR value indicates a lower difference between the two images, and therefore a higher similarity.
    \item Structural Similarity Index (SSIM) measures the similarity of two images by comparing their luminance, contrast, and structural information. A higher SSIM value indicates a higher similarity between the two images.
    \item Learned Perceptual Image Patch Similarity (LPIPS) measures the similarity between two images by comparing their perceptual features learned from a pre-trained deep neural network. A lower LPIPS value indicates a higher similarity between the two images.
\end{itemize}
\subsection{Results}

In this study, we conducted a comprehensive evaluation of our proposed hybrid framework, BRADCN, for synthetic bokeh rendering, as well as its performance against state-of-the-art models. Specifically, we compared both the quantitative and qualitative results of our model against other state-of-the-art models in the Val294 set, which demonstrated the superiority of our model in producing a more effective bokeh effect.

\begin{table}[htbp]
\centering
\caption{Quantitative results on Val294, comparing our proposed BRADCN framework with state-of-the-art methods in terms of PSNR, SSIM and LPIPS metrics.}\label{tab1}
\setlength{\tabcolsep}{5mm}{
\begin{tabular}{|c|c|c|c|}
\hline
\textbf{Method} & \textbf{PSNR}$\uparrow$ & \textbf{SSIM}$\uparrow$ & \textbf{LPIPS}$\downarrow$\\
 \hline
 SKN \cite{li2019selective} & 24.66 & 0.8521 & 0.3323 \\ 
 DBSI \cite{dutta2021depth} & 23.45 & 0.8675 & 0.24633 \\ 
 DMSHN \cite{dutta2021stacked} & 24.72 & 0.8793 & 0.2271\\ 
 BRVIT \cite{nagasubramaniam2022bokeh} & \underline{24.76} & \textbf{0.8904} & \underline{0.1924}\\
 \hline
 Ours (BRADCN) & \textbf{24.83} & \underline{0.8737} & \textbf{0.1448}\\
 \hline
\end{tabular}
}
\end{table}
To evaluate the performance of our model, we compared several objective evaluation metrics, including the LPIPS, SSIM and PSNR. The results showed that our model achieved a remarkable \textbf{24.74}\% improvement on LPIPS and obtained a higher PSNR compared to the other state-of-the-art models. A detailed summary of all the quantitative results for comparison is presented in Table \ref{tab1}.


To further evaluate the effectiveness of our model in generating a realistic bokeh effect, we conducted a visual effect comparison between our model and the other models, as illustrated in Fig. \ref{V}. The results clearly demonstrate that our model generates a more stable and realistic bokeh effect compared to other models. Additionally, our model produces a more natural and smoother transition between the foreground and background, leading to a clearer and more focused foreground.

\begin{figure}[htbp]
\centerline{\includegraphics[width=9cm]{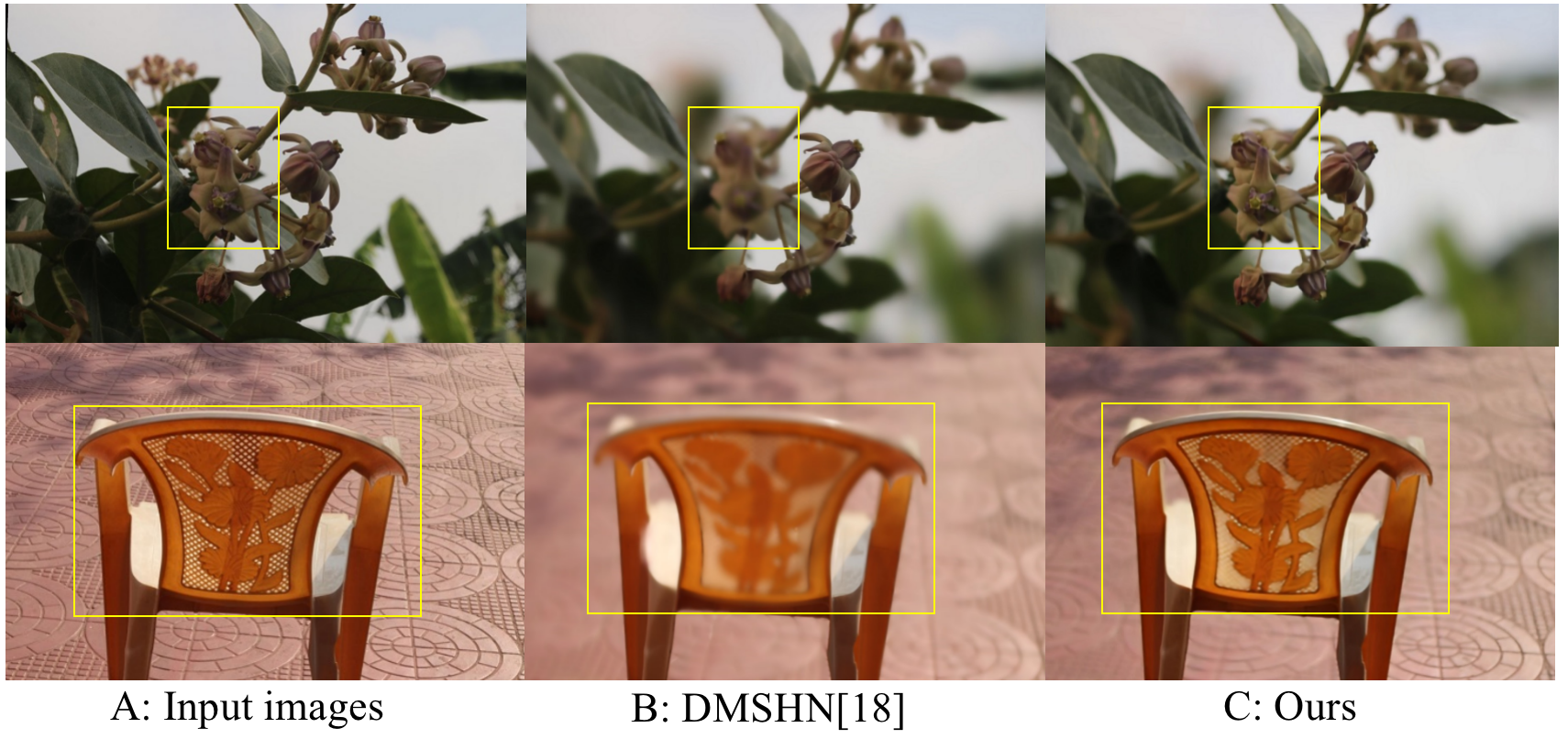}}
\caption{Qualitative results comparison on EBB! Test set. Our proposed BRADCN framework is able to achieve a more stable and realistic bokeh effect compared to other state-of-the-art methods.}
\label{T}
\end{figure}

In addition to the Val294 set, we also evaluated the generalization of our model on the EBB Test set. Since the EBB Test set does not provide a corresponding bokeh effect image, we compared the results with the state-of-the-art model\cite{dutta2021stacked} in Fig. \ref{T}. The results showed that our model outperformed the state-of-the-art model in terms of visual quality and overall effectiveness.

\subsection{Ablation Study}

To assess the effectiveness of each component in our model, we conducted a series of ablation experiments. The results of these experiments are presented in Table \ref{tab2}. We found that the inclusion of depth information in the rendering process leads to a slight improvement in the rendering quality, which is reflected in the increased values of PSNR and SSIM metrics. Moreover, the addition of ADCN to the model further enhances the accuracy of the depth information, thereby increasing the stability and realism of the bokeh rendering effect. This is evident in the significant improvement in the objective evaluation metrics, such as LPIPS, compared to the model without ADCN. These results demonstrate the effectiveness of ADCN in adjusting the errors in depth prediction, which is crucial for the stability and quality of the final rendering output. Overall, the ablation experiments confirm the validity of our model design and the contribution of each sub-model to the final rendering output.
\begin{table}[h]
\centering
\caption{Ablation study metrics on Val294. The table shows the results of different ablation settings on the Val294 dataset using the proposed framework. The best results are highlighted in bold.}\label{tab2}
\setlength{\tabcolsep}{3.8mm}{
\begin{tabular}{|c|c|c|c|}
\hline
\textbf{Method} & \textbf{PSNR}$\uparrow$ & \textbf{SSIM}$\uparrow$ & \textbf{LPIPS}$\downarrow$\\
 \hline
 Render Net (only) & 24.29 & 0.8671 & 0.1530 \\ 
 Render Net with depth & 23.43 & 0.8688 & 0.1529 \\ 
 \hline
 Ours (BRADCN) & \textbf{24.83} & \textbf{0.8737} & \textbf{0.1448}\\
 \hline
\end{tabular}
}
\end{table}

\subsection{Experiment Setting}

The implementation of our model is conducted on the 32 GB NVIDIA Tesla V100 GPUs. To ensure a stable training process, we set the batch size to 1. During the training stage, the input resolution of the images is set to 768 × 512. After the training is completed, we use bilinear interpolation to double the output resolution for testing. This allows us to compare the results with the ground truth images and evaluate the performance of the model in generating high-quality bokeh effects.

\section{CONCLUSION}
In this paper, we present a new hybrid model for bokeh rendering, which incorporates the Adaptive Depth Correction Network (ADCN) to address inaccuracies in monocular depth estimation. Our approach achieves a more stable, realistic, and natural rendering effect compared to state-of-the-art methods. We evaluate our model using various objective quality metrics such as PSNR, SSIM, and LPIPS, and demonstrate its superiority in bokeh effect generation. The proposed model is logically designed to adaptively calibrate the depth information, which is critical to generate the desired bokeh effect. Through our experiments, we demonstrate the effectiveness of the ADCN in improving the stability of the bokeh rendering effect, and how it significantly reduces the error in depth prediction. By combining L1 loss function and MS-SSIM loss function, our model is able to achieve a higher level of performance in terms of objective metrics.

In future work, we plan to optimize our model by incorporating semantic segmentation to extract the main semantics from the input image, which would be useful for a range of tasks. This can be achieved through multi-task learning, which has been shown to be an effective way to improve the performance of deep learning models. Our hybrid model is a significant contribution to the field of bokeh rendering, and we believe it has the potential to be further improved and applied in various computer vision tasks.

\bibliographystyle{IEEEtran}
\bibliography{ICCCS/BRADCN}

\begin{thebibliography}{10}
\providecommand{\url}[1]{#1}
\csname url@samestyle\endcsname
\providecommand{\newblock}{\relax}
\providecommand{\bibinfo}[2]{#2}
\providecommand{\BIBentrySTDinterwordspacing}{\spaceskip=0pt\relax}
\providecommand{\BIBentryALTinterwordstretchfactor}{4}
\providecommand{\BIBentryALTinterwordspacing}{\spaceskip=\fontdimen2\font plus
\BIBentryALTinterwordstretchfactor\fontdimen3\font minus
  \fontdimen4\font\relax}
\providecommand{\BIBforeignlanguage}[2]{{%
\expandafter\ifx\csname l@#1\endcsname\relax
\typeout{** WARNING: IEEEtran.bst: No hyphenation pattern has been}%
\typeout{** loaded for the language `#1'. Using the pattern for}%
\typeout{** the default language instead.}%
\else
\language=\csname l@#1\endcsname
\fi
#2}}
\providecommand{\BIBdecl}{\relax}
\BIBdecl

\bibitem{shen2016automatic}
X.~Shen, A.~Hertzmann, J.~Jia, S.~Paris, B.~Price, E.~Shechtman, and I.~Sachs,
  ``Automatic portrait segmentation for image stylization,'' in \emph{Computer
  Graphics Forum}, vol.~35.\hskip 1em plus 0.5em minus 0.4em\relax Wiley Online
  Library, 2016, pp. 93--102.

\bibitem{shen2016deep}
X.~Shen, X.~Tao, H.~Gao, C.~Zhou, and J.~Jia, ``Deep automatic portrait
  matting,'' in \emph{Proceedings of the European Conference on Computer
  Vision}.\hskip 1em plus 0.5em minus 0.4em\relax Springer, 2016, pp. 92--107.

\bibitem{ignatov2020rendering}
A.~Ignatov, J.~Patel, and R.~Timofte, ``Rendering natural camera bokeh effect
  with deep learning,'' in \emph{Proceedings of the IEEE/CVF Conference on
  Computer Vision and Pattern Recognition Workshops}, 2020, pp. 418--419.

\bibitem{nagasubramaniam2022bokeh}
H.~Nagasubramaniam and R.~Younes, ``Bokeh effect rendering with vision
  transformers,'' \emph{TechRxiv Preprint techrxiv:17714849.v1}, 2022.

\bibitem{lee2022bokeh}
B.~Lee, F.~Lei, H.~Chen, and A.~Baudron, ``Bokeh-loss gan: Multi-stage
  adversarial training for realistic edge-aware bokeh,'' \emph{arXiv preprint
  arXiv:2208.12343}, 2022.

\bibitem{dosovitskiy2020image}
A.~Dosovitskiy, L.~Beyer, A.~Kolesnikov, D.~Weissenborn, X.~Zhai,
  T.~Unterthiner, M.~Dehghani, M.~Minderer, G.~Heigold, S.~Gelly \emph{et~al.},
  ``An image is worth 16x16 words: Transformers for image recognition at
  scale,'' \emph{arXiv preprint arXiv:2010.11929}, 2020.

\bibitem{ranftl2021vision}
R.~Ranftl, A.~Bochkovskiy, and V.~Koltun, ``Vision transformers for dense
  prediction,'' in \emph{Proceedings of the IEEE/CVF International Conference
  on Computer Vision}, 2021, pp. 12\,179--12\,188.

\bibitem{abadie2018advances}
G.~Abadie, S.~McAuley, E.~Golubev, S.~Hill, and S.~Lagarde, ``Advances in
  real-time rendering in games,'' in \emph{ACM SIGGRAPH 2018 Courses}.\hskip
  1em plus 0.5em minus 0.4em\relax ACM SIGGRAPH 2018 Courses, 2018, pp. 1--1.

\bibitem{luo2020bokeh}
X.~Luo, J.~Peng, K.~Xian, Z.~Wu, and Z.~Cao, ``Bokeh rendering from defocus
  estimation,'' in \emph{Proceedings of the European Conference on Computer
  Vision}.\hskip 1em plus 0.5em minus 0.4em\relax Springer, 2020, pp. 245--261.

\bibitem{lijun2018deeplens}
W.~Lijun, S.~Xiaohui, Z.~Jianming, W.~Oliver, H.~Chih-Yao, K.~Sarah, and
  L.~Huchuan, ``Deeplens: Shallow depth of field from a single image,'' in
  \emph{ACM Trans. Graph.(Proc. SIGGRAPH Asia)}, vol.~37, 2018, pp. 1--11.

\bibitem{peng2022bokehme}
J.~Peng, Z.~Cao, X.~Luo, H.~Lu, K.~Xian, and J.~Zhang, ``Bokehme: When neural
  rendering meets classical rendering,'' in \emph{Proceedings of the IEEE/CVF
  Conference on Computer Vision and Pattern Recognition}, 2022, pp.
  16\,283--16\,292.

\bibitem{song2015sun}
S.~Song, S.~P. Lichtenberg, and J.~Xiao, ``Sun rgb-d: A rgb-d scene
  understanding benchmark suite,'' in \emph{Proceedings of the IEEE conference
  on computer vision and pattern recognition}, 2015, pp. 567--576.

\bibitem{ronneberger2015u}
O.~Ronneberger, P.~Fischer, and T.~Brox, ``U-net: Convolutional networks for
  biomedical image segmentation,'' in \emph{Proceedings of the International
  Conference on Medical image computing and computer-assisted
  intervention}.\hskip 1em plus 0.5em minus 0.4em\relax Springer, 2015, pp.
  234--241.

\bibitem{zhao2016loss}
H.~Zhao, O.~Gallo, I.~Frosio, and J.~Kautz, ``Loss functions for image
  restoration with neural networks,'' \emph{IEEE Transactions on computational
  imaging}, vol.~3, no.~1, pp. 47--57, 2016.

\bibitem{wang2003multiscale}
Z.~Wang, E.~P. Simoncelli, and A.~C. Bovik, ``Multiscale structural similarity
  for image quality assessment,'' in \emph{Proceedings of the The
  Thrity-Seventh Asilomar Conference on Signals, Systems \& Computers, 2003},
  vol.~2.\hskip 1em plus 0.5em minus 0.4em\relax Ieee, 2003, pp. 1398--1402.

\bibitem{paul2019analysis}
E.~S. Paul and J.~Anitha, ``Analysis of transform-based compression techniques
  for mri and ct images,'' in \emph{Intelligent Data Analysis for Biomedical
  Applications}.\hskip 1em plus 0.5em minus 0.4em\relax Elsevier, 2019, pp.
  103--120.

\bibitem{zhang2018unreasonable}
R.~Zhang, P.~Isola, A.~A. Efros, E.~Shechtman, and O.~Wang, ``The unreasonable
  effectiveness of deep features as a perceptual metric,'' in \emph{Proceedings
  of the IEEE conference on computer vision and pattern recognition}, 2018, pp.
  586--595.

\bibitem{li2019selective}
X.~Li, W.~Wang, X.~Hu, and J.~Yang, ``Selective kernel networks,'' in
  \emph{Proceedings of the IEEE/CVF conference on computer vision and pattern
  recognition}, 2019, pp. 510--519.

\bibitem{dutta2021depth}
S.~Dutta, ``Depth-aware blending of smoothed images for bokeh effect
  generation,'' \emph{Journal of Visual Communication and Image
  Representation}, vol.~77, p. 103089, 2021.

\bibitem{dutta2021stacked}
S.~Dutta, S.~D. Das, N.~A. Shah, and A.~K. Tiwari, ``Stacked deep multi-scale
  hierarchical network for fast bokeh effect rendering from a single image,''
  in \emph{Proceedings of the IEEE/CVF Conference on Computer Vision and
  Pattern Recognition}, 2021, pp. 2398--2407.

\end{thebibliography}


 \end{document}